\documentclass[12pt]{article}
\usepackage{subcaption}
\usepackage[bottom]{footmisc}
\usepackage{indentfirst} 
\usepackage{stmaryrd}
\usepackage{endnotes}
\usepackage{verbatim}
\usepackage{tikz}
\usepackage{threeparttable}
\usepackage{comment}
\usepackage{titlesec}
\usepackage{multirow}
\usepackage{makecell}
\usepackage{titling}
\usepackage{chngcntr}
\usepackage{dsfont}
\usepackage{hyperref}
\usepackage{enumitem}
\usepackage{graphicx,float}
\usepackage{amsmath,amsthm,amssymb,color}
\usepackage[utf8]{inputenc} 
\usepackage[T1]{fontenc}    
\usepackage{hyperref}       
\usepackage{url}            
\usepackage{booktabs}       
\usepackage{amsfonts}       
\usepackage{nicefrac}       
\usepackage{microtype}      
\usepackage{xcolor}         
\usepackage{titlesec}
\usepackage{setspace}
\usepackage{array}
\usepackage[titletoc]{appendix}
\usepackage{natbib}
\usepackage{geometry}
\bibpunct{(}{)}{;}{a}{,}{,}
\newcommand{\bff}{\mathrm{\bf f}}
\newcommand{\bx}{\mathrm{\bf x}}

\usepackage{amsmath,amsthm,amssymb}
\usepackage{tabulary}
\usepackage{colortbl}
\usepackage{graphicx}       
\usepackage{subcaption}
\usepackage[bottom]{footmisc}
\usepackage{indentfirst} 
\usepackage{stmaryrd}
\usepackage{endnotes}
\usepackage{verbatim}
\usepackage{tikz}
\usepackage{threeparttable}
\usepackage{comment}
\usepackage{algorithm}
\usepackage{algpseudocode}
\usepackage{titlesec}
\usepackage{multirow}
\usepackage{makecell}

\allowdisplaybreaks[4]
\begin{document}

\newcounter{appendixcounter}
\renewcommand{\theappendixcounter}{\Alph{appendixcounter}}
\refstepcounter{appendixcounter}
\label{append:staggered-design}
\refstepcounter{appendixcounter}
\label{section:proofs}
\refstepcounter{appendixcounter}
\label{append:simulation}
\refstepcounter{appendixcounter}
\label{append: covid19}

\title{Factor Augmented Supervised Learning with Text
Embeddings}

\author{
Zhanye Luo$^1$, Yuefeng Han$^2$, and Xiufan Yu$^2$ \\ $^1$University of Chicago, $^2$University of Notre Dame
}
\date{}
\maketitle{}

\pagestyle{plain}

\setstretch{1.5}
\begin{abstract}
Large language models (LLMs) generate text embeddings from text data, producing vector representations that capture the semantic meaning and contextual relationships of words. However, the high dimensionality of these embeddings often impedes efficiency and drives up computational cost in downstream tasks. To address this, we propose AutoEncoder‑Augmented Learning with Text (AEALT), a supervised, factor‑augmented framework that incorporates dimension reduction directly into pre‑trained LLM workflows. First, we extract embeddings from text documents; next, we pass them through a supervised augmented autoencoder to learn low‑dimensional, task‑relevant latent factors. By modeling the nonlinear structure of complex embeddings, AEALT outperforms conventional deep‑learning approaches that rely on raw embeddings. We validate its broad applicability with extensive experiments on classification, anomaly detection, and prediction tasks using multiple real‑world public datasets. Numerical results demonstrate that AEALT yields substantial gains over both vanilla embeddings and several standard dimension reduction methods.
\end{abstract}

\noindent Keywords: Dimension reduction, High-dimensional embeddings, Large language model, Nonlinear factor model, Supervised autoencoders

\newpage
\setstretch{1.5}
\section{Introduction}

Text analysis has become increasingly popular across various domains, including finance \citep{li2023large,nie2024survey}, economics \citep{bybee2024business,cong2024textual}, healthcare \citep{cascella2023evaluating,peng2023study}, marketing \citep{brand2023using,arora2025ai}, and social science \citep{grossmann2023ai,manning2024automated}, among others. 
The advances of large language models (LLMs) have significantly enhanced the capacity to leverage text data as predictive features in different learning tasks.
For example, financial reports and earnings call transcripts can be used to forecast stock market movements by extracting sentiment, forward-looking statements, and key financial indicators from the text \citep{lopez2023can};
customer reviews and query logs can be used to predict product demand in e-commerce by capturing trends in consumer sentiment, interest, and purchasing intent over time \citep{zhang2022forecasting};
transaction descriptions can be used in fraud detection to flag suspicious activities by identifying unusual patterns, keywords, or behaviors indicative of fraudulent behavior \citep{boulieris2024fraud}. 

Heuristic approaches frequently transform text inputs into high-dimensional vector-valued representations (also known as \emph{text embeddings}), and then directly utilize these high-dimensional embeddings as input features for downstream learning tasks \citep{kim2014convolutional,li2023large,nie2024survey,su2024large}.
While this strategy simplifies the pipeline by reducing the need for manual feature engineering, it also presents several limitations.
\emph{First}, the high dimensionality of embeddings often introduces redundancy and multicollinearity, which can lead to overfitting and obscure task-relevant signals. If the dependencies among embedding dimensions are not properly accounted for, this could compromise both the effectiveness of data representation and the interpretability of the resulting models \citep{mu2018allbutthetop,drinkall2025dimensionality}. 
\emph{Second}, the high dimensionality of LLM embeddings demands substantial computational resources and memory usage during both model training and model fitting. Large embedding vectors demand extensive storage memory and slow down optimization, raising the cost of model deployment \citep{ling-etal-2016-word}.

In this article, we propose the \textbf{AutoEncoder Augmented Learning with Text (AEALT)}, a unifying framework for supervised learning tasks with textual input. Different from heuristic methods that directly feed high-dimensional embeddings into downstream learning algorithms, AEALT introduces a (possibly) non-linear factor model \citep{yalcin2001nonlinear,xiu2024deep} to first extract informative latent factors from the embeddings prior to model fitting. This factor analysis is realized via a supervised autoencoder \citep{le2018supervised} that optimizes a composite loss that balances reconstruction fidelity with predictive accuracy, ensuring that the extracted latent representations are both compact and well-aligned with downstream objectives. Our contributions can be summarized in three folds. 
\begin{itemize}[leftmargin = 5ex]
    \item  AEALT offers a general framework for supervised learning tasks involving textual input. It is not confined to a specific task, but rather offers flexibility and adaptivity to a wide range of applications, including classification, prediction, and beyond.
    In our numerical studies, we demonstrate the practicality of AEALT through three real-world applications: sentiment analysis, anomaly detection, and price prediction. 
\end{itemize} 
\begin{itemize}[leftmargin = 5ex]
    \item AEALT explicitly incorporates supervised dimension reduction by integrating the target variable into the factor extraction process through a supervised autoencoder.
    This approach utilizes the representational power of deep neural network architectures to efficiently capture complex nonlinear relationships from high-dimensional text embeddings.  
    Empirical studies show that AEALT delivers strong predictive performance across different tasks, consistently outperforming both the benchmark without factor extraction and alternative methods based on unsupervised dimension reduction. 
\end{itemize} 
\begin{itemize}[leftmargin = 5ex]
    \item  AEALT provides a unifying modeling pipeline that integrates a factor-structured model to extract key predictive latent factors from high-dimensional embeddings. By allowing for different factor loading functions, AEALT unifies a number of factor analysis approaches, such as principal component analysis (PCA), autoencoders (AE), linear and nonlinear factor models. 
\end{itemize} 

The paper is organized as follows. Section \ref{sec:related-works} discusses related works in the literature.
Section \ref{sec:methodology} introduces methodological details of the proposal AEALT. 
Section \ref{sec:evaluation} presents extensive empirical results to demonstrate the practicability of the proposed method in real-world use cases. 
Section \ref{sec:conclusion} concludes the paper with a brief discussion.

\section{Related Works} \label{sec:related-works}

\paragraph{Post-processing for Text Embeddings.}
Text embeddings, such as those produced by Word2Vec \citep{mikolov2013efficient}, GloVe \citep{pennington2014glove}, BERT \citep{devlin-etal-2019-bert}, LLM embedding models \citep{openai2022embedding}, and their successors, often reside in high-dimensional vector spaces to capture rich semantic and syntactic properties. 
However, these high-dimensional embeddings can be computationally expensive and may contain redundant components that hinder downstream performance. As a result, researchers started exploring the impact of dimensionality \citep{yin2018dimensionality,drinkall2025dimensionality} and dimension reduction techniques \citep{mu2018allbutthetop,ke2022svd,hwang2023embedtextnet,zhou2024machines} to improve the efficiency and interpretability of text representations. 
Among various approaches, PCA has stood out due to its simplicity and computational efficiency \citep{wang2019single,zhang2024evaluating,khaledian2025pcarag}. However, PCA is limited by its linear nature, which may fail to capture the complex semantic relationships often present in high-dimensional text embeddings. 
One strategy to alleviate the linearity restriction is to employ AE models \citep{zhang2024evaluating,drinkall2025dimensionality}, which leverage nonlinear transformations to learn more expressive lower-dimensional embeddings. Most of these efforts have predominantly focused on unsupervised dimension reduction approaches, emphasizing the preservation of semantic consistency and the improvement of computing efficiency, while giving limited consideration to how the reduced representations interface with downstream tasks. 
This motivates our research on the development of task-oriented dimension reduction methods for extracting task-relevant representations from text embeddings.

\paragraph{Factor Models for Dimension Reduction.} 
Factor models \citep{lawley1962factor,Yalcin2001NonlinearFA,bai2002determining,bai2003inferential} have long served as a foundational tool for dimension reduction in the statistical analysis of high-dimensional data. They achieve dimension reduction by capturing the underlying structure of the data through a smaller set of unobserved latent variables (a.k.a., factors) which are assumed to explain the shared variation among observed variables. In time series forecasting involving a large number of predictors, factor models are often used as a preliminary step to extract informative low-dimensional representations from high-dimensional inputs \citep{stock2002forecasting,bair2006prediction,fan2017sufficient,yu2022nonparametric,luo2022inverse,zhou2025factor}. These representations can then be fed into suitable learning algorithms to improve efficiency, interpretability, and predictive performance. Our research in this article is motivated by the ability of factor models to distill high-dimensional signals into concise latent structures that preserve predictive information. Conventional factor models typically extract latent factors in an unsupervised manner, focusing solely on capturing the common variation among predictors, which may lead to the overlook of features relevant to the response variable. To overcome this limitation, in our proposed method, we extract the latent factors in a supervised manner, aiming to recover latent representations that are not only low-dimensional but also more informative for downstream predictive tasks.

\section{Methodology} \label{sec:methodology}

\begin{figure}[H]
\centering
\includegraphics[width=0.96\linewidth]{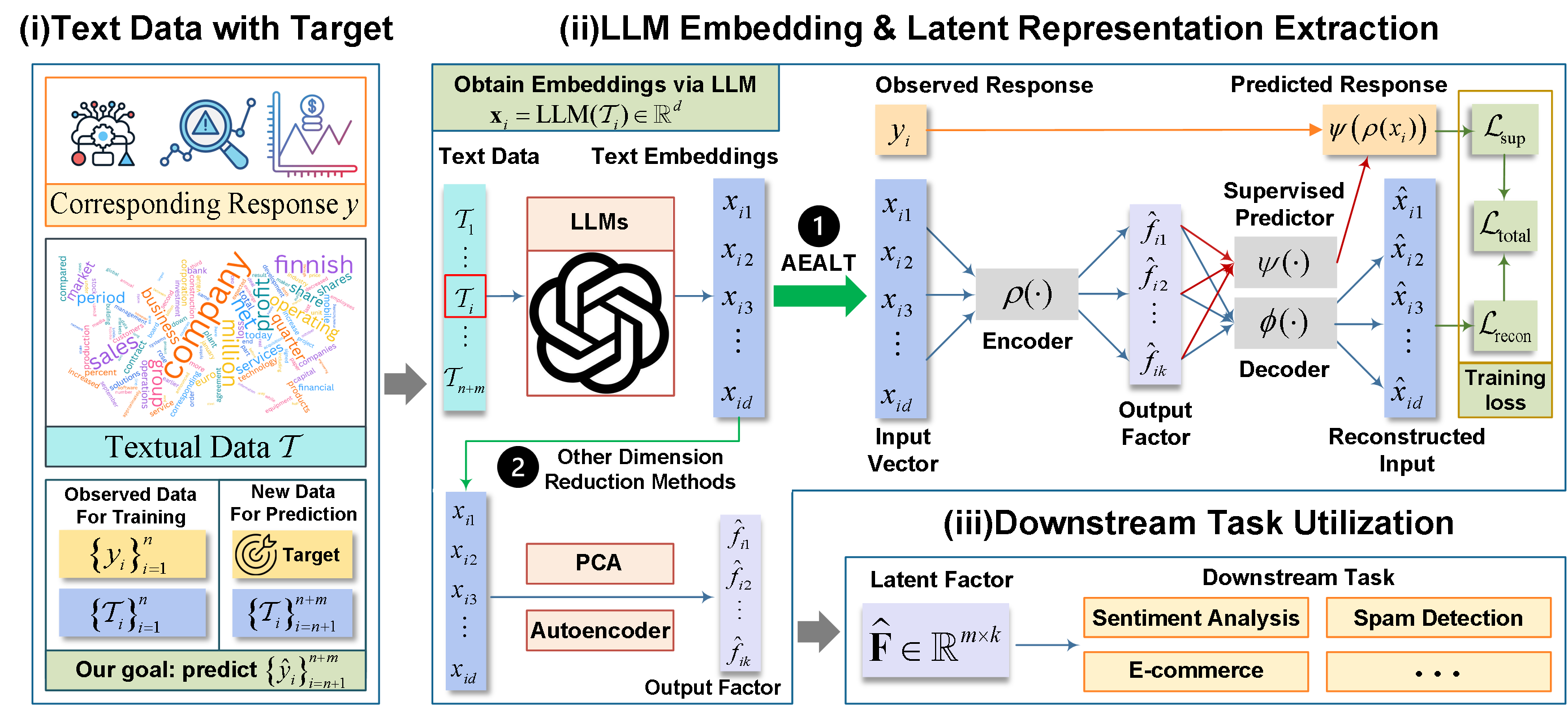}
\caption{ A graphical illustration of the proposed AutoEncoder-Augmented Learning with Text (AEALT) framework. 
The pipeline consists of three stages: (i) Collecting observed text data and analyze targets as $\mathcal{D} = \{(\mathcal{T}_i, y_i)\}_{i=1}^{n}$ as well as newly observed samples \( \left\{ \mathcal{T}_i \right\}_{i = n+1}^{n+m} \); 
(ii) Obtain textual embeddings \( \bx_i \in \mathbb{R}^d \) via pre-trained LLMs, followed by dimension reduction through AEALT, jointly learning reconstruction and predictive objectives or alternative methods (e.g., AE and PCA); 
(iii) Conduct downstream analysis by estimating the target values \( \{ \hat{y}_i \}_{i = n+1}^{n+m} \) using the representations $\{ \widehat{\boldsymbol{f}}_i \}_{i = n+1}^{n+m}$. 
}
\label{fig:workflow}
\end{figure}

In this section, we describe a complete pipeline from raw text data to downstream analysis. We begin by converting each text document into a high-dimensional embedding using a pre-trained LLM, then apply dimension reduction to extract task-relevant, low-dimensional latent representations (factors). We introduce \textbf{AutoEncoder-Augmented Learning with Text (AEALT)}, a supervised framework that jointly reconstructs embeddings and aligns them with task-specific objectives, alongside standard techniques such as PCA and vanilla autoencoders. Finally, the resulting low-dimensional latent representations are fed into models for downstream tasks such as classification, prediction, or anomaly detection. The overall workflow is illustrated in Figure~\ref{fig:workflow}, and AEALT's details appear in Algorithm~\ref{alg:AEALT}.

\subsection{Problem Setup} 
Suppose we have a text-based dataset of $n$ samples $\mathcal{D} = \{(\mathcal{T}_i, y_i)\}_{i=1}^n$, where $\mathcal{T}_i$ is the $i$-th text document and $y_i$ is the associated target response, and $n$ is the sample size of observed data. Here, $y_i$ can take various data forms depending on the learning tasks, e.g.,  categorical labels for classification, continuous values for regression, or structured outputs for sequence labeling tasks. For some newly observed text documents $\{ \mathcal{T}_i \}_{i=n+1}^{n+m}$, \emph{the prediction task in text learning is to accurately predict the corresponding responses $\{y_i\}_{i=n+1}^{n+m}$, given the observed training data $\{ (\mathcal{T}_i, y_i) \}_{i=1}^{n}$ and the new text documents $\{ \mathcal{T}_i \}_{i=n+1}^{n+m}$}. Here, $m$ is the number of new samples to predict.

\begin{algorithm}[H]
\caption{AEALT: AutoEncoder Augmented Learning with Text}
\label{alg:AEALT}
\begin{algorithmic}[1]
\Require Training set: raw text data $\{ \mathcal{T}_i \}_{i=1}^n$, target variables $\{y_i\}_{i=1}^n$; testing set: raw text data $\{ \mathcal{T}_i \}_{i=n+1}^{n+m}$.
\Ensure Predictions for testing set target variables $\{\hat{y}_i\}_{i=n+1}^{n+m}$.

\State \textbf{Step 1: Text Embeddings}  
\For{training and testing sample \( i = 1, \dots, n+m \)}
    \State Use pre‑trained LLMs to embed text document $\mathcal{T}_i$ to obtain an embedding vector $\mathbf{x}_i \in \mathbb{R}^d$ as the representational features of the text document, i.e., $\mathbf{x}_i = \mathrm{LLM}(\mathcal{T}_i) \in \mathbb{R}^{d}$.

\EndFor

\State \textbf{Step 2: Latent Representation Learning via Supervised Autoencoder}  

\State Train a supervised autoencoder using loss function
\begin{align*}
&\hat{\Theta}_e, \hat{\Theta}_d, \hat{\Theta}_p 
= \arg\min_{\Theta_e, \Theta_d, \Theta_p} 
\left\{ (1 - \lambda)\, \mathcal{L}_{\text{recon}} 
+ \lambda\, \mathcal{L}_{\text{sup}} \right\}   ,
\end{align*}
where the reconstruction loss $\mathcal{L}_{\text{recon}}$ and supervised loss $\mathcal{L}_{\text{sup}}$ are defined as
\begin{align*}
&\mathcal{L}_{\text{recon}}=\sum_{i=1}^n \left\| \varphi(\rho(\mathbf{x}_i;\Theta_e); \Theta_d) - \mathbf{x}_i \right\|^2 , \\
&\mathcal{L}_{\text{sup}}=\sum_{i=1}^n \mathcal R\left( \psi(\rho(\mathbf{x}_i;\Theta_e); \Theta_p), y_i \right) ,
\end{align*}
and $\mathcal R$ is a task-specific loss function for each downstream task.

\For{training and testing sample \( i = 1, \dots, n+m \)}
    \State Compute the latent representations using the trained supervised autoencoder $\hat\bff_i = \rho(\mathbf{x}_i; \hat\Theta_e)$.
\EndFor

\State \textbf{Step 3: Downstream Learning (classification, anomaly detection, regression, etc.)}  

\State Fit a neural network or classical model $\mathcal{H}(\cdot;\Theta_t)$ on the training data $\{\hat\bff_i, y_i,i=1,\dots, n\}$,
\begin{align*}
\hat \Theta_t=\arg\min_{\Theta_t} \sum_{i=1}^n \mathcal R (\mathcal H( \hat\bff_i;\Theta_t),y_i).  
\end{align*}
where the loss function $\mathcal R(\cdot,\cdot)$ is problem- and algorithm-specific.

\For{testing sample \( i = n+1, \dots, n+m \)}
    \State Predict $\hat y_i=\mathcal{H}(\hat\bff_i;\hat\Theta_t)$.
\EndFor

\end{algorithmic}
\end{algorithm}

\subsection{Learning Latent Representation from LLM Embeddings}

\paragraph{Text embeddings.}
Each text document $\mathcal{T}_i$ can be transformed by LLMs into text embeddings $\bx_i$,
\begin{equation}
\mathbf{x}_i = \mathrm{LLM}(\mathcal{T}_i) \in \mathbb{R}^{d},
\label{eq:llm}
\end{equation}
where $\bx_i$ is a $d$ dimensional vector, and $d$ represents the dimensionality of the embedding space. These embeddings provide two key advantages. First, they enhance topic coherence. When incorporated into neural topic models, contextualized document embeddings produce substantially more coherent and interpretable topics than traditional bag-of-words approaches \cite{bianchi-etal-2021-pre}.
Second, they capture semantic relationships between words. For example, LLMs tend to cluster semantically related words more tightly than classical models \cite{freestone2024word}. Thanks to these benefits, the text embeddings generated by LLMs encapsulate richer information than the word count data traditionally used in text analysis.

\paragraph{Latent representation learning.}
LLM embeddings often suffer from high dimensionality, which can undermine statistical efficiency and increase computational cost. For example, the embedding dimension of the original BERT model, specifically the BERT base model, is 768. To address this, we investigate several dimension reduction techniques as illustrated in Figure~\ref{fig:workflow}, including our proposed AEALT, which explicitly aligns the resulting low‑dimensional representations with downstream tasks, and standard methods such as principal component analysis (PCA) and vanilla autoencoders.

By incorporating predictive tasks into the autoencoder, AEALT exploits a shared low‑dimensional representation structure to enhance both embedding reconstruction and downstream prediction, yielding a more versatile and powerful model for applications that benefit from jointly modeling inputs and outputs.

We define the structure of AEALT (after embedding) as follows: for each text embedding vector \( \mathbf{x}_i \in \mathbb{R}^d \), we employ a supervised autoencoder comprising three neural network modules: 

\begin{enumerate}
    \item[(i)] {an encoder function} \( \rho(\cdot; \Theta_e) \), 
    \item[(ii)] {a decoder function} \( \varphi(\cdot; \Theta_d) \), 
    \item[(iii)] {a supervised prediction network} \( \psi(\cdot; \Theta_p) \). 
\end{enumerate}

The parameters of these neural networks are defined as \( \Theta_e = \{W^{(e)}_l, \mathbf{b}^{(e)}_l\}_{l=1}^{L} \), \( \Theta_d = \{W^{(d)}_l, \mathbf{b}^{(d)}_l\}_{l=1}^{L} \), and \( \Theta_p = \{W^{(p)}_l, \mathbf{b}^{(p)}_l\}_{l=1}^{L_p} \), where \( L \) is the number of layers in both the encoder and decoder, and \( L_p \) is the number of layers in the prediction network. For each input text embedding $\bx_i$, these modules produce the estimated latent representation $\bff_i^*=\rho(\mathbf{x}_i; \Theta_e)$, reconstructed input embedding $\varphi(\bff_i^* ; \Theta_d)$ and predicted response $\psi(\bff_i^* ; \Theta_p)$,
\begin{align*}
&\bff_i^*=\rho(\mathbf{x}_i; \Theta_e) = \sigma_{L} \left( W^{(e)}_{L} \cdots \sigma_1 \left( W^{(e)}_1 \mathbf{x}_i + \mathbf{b}^{(e)}_1 \right) + \cdots + \mathbf{b}^{(e)}_{L} \right), \\
& \varphi(\bff_i^* ; \Theta_d) = \sigma_{L} \left( W^{(d)}_{L} \cdots \sigma_1 \left( W^{(d)}_1 \bff_i^* + \mathbf{b}^{(d)}_1 \right) + \cdots + \mathbf{b}^{(d)}_{L} \right), \\
& \psi(\bff_i^* ; \Theta_p) = \sigma_{L} \left( W^{(p)}_{L} \cdots \sigma_1 \left( W^{(p)}_1 \bff_i^* + \mathbf{b}^{(p)}_1 \right) + \cdots + \mathbf{b}^{(p)}_{L} \right) ,
\end{align*}
where each $\sigma_l$ denotes the activation function at layer $l$. This framework generalizes the standard autoencoder by jointly reconstructing $\bx_i$ and predicting $y_i$; it can be viewed as an autoencoder operating on the joint space $(\bx_i,y_i)$ with the encoder weights corresponding to $y_i$ held fixed at zero.

The neural network parameters are learned by jointly minimizing reconstruction and supervised losses:
\begin{align*}
&\hat{\Theta}_e, \hat{\Theta}_d, \hat{\Theta}_p 
= \arg\min_{\Theta_e, \Theta_d, \Theta_p} 
\left\{ (1 - \lambda)\, \mathcal{L}_{\text{recon}} 
+ \lambda\, \mathcal{L}_{\text{sup}} \right\}   ,
\end{align*}
where
\begin{align*}
\vspace{-3ex}
&\mathcal{L}_{\text{recon}}=\sum_{i=1}^n \left\| \varphi(\rho(\mathbf{x}_i;\Theta_e); \Theta_d) - \mathbf{x}_i \right\|^2 , \\
&\mathcal{L}_{\text{sup}}=\sum_{i=1}^n \mathcal R\left( \psi(\rho(\mathbf{x}_i;\Theta_e); \Theta_p), y_i \right) ,
\end{align*}
where $\mathcal{L}_{\text{recon}}$ is the reconstruction error, $\mathcal{L}_{\text{sup}}$ is the supervised loss, and the weight $\lambda \in [0, 1]$ is a hyperparameter that controls the trade-off between the two loss functions. 

\paragraph{Task-dependent supervised loss. } $\mathcal{L}_{\text{sup}}$ is a supervised loss function whose explicit form depends on the supervised learning task of interest. Such a task-dependent loss function enables the generality of the proposed AEALT, and offers flexibility and adaptivity to a wide range of applications. 
Depending on the downstream task, $\mathcal{L}_{\text{sup}}$ may be a classification loss (e.g., cross‑entropy) or a regression loss (e.g., mean squared error). In particular, the default choice for classification and anomaly detection is the cross-entropy loss, defined as $\mathcal R(\hat y_i, y_i) = - \sum_{j=1}^c y_{ij} \log (\hat{y}_{ij}),$
where $c$ is the number of classes and $y_i=(y_{i1},\ldots,y_{ic})^\top$. Meanwhile, the default choice for regression is the mean squared error, defined as $\mathcal R(\hat y_i, y_i)= \| \hat{y}_i - y_i \|^2.$
The resulting task‑relevant latent representations (or factors) are $\hat \bff_i:= \rho(\mathbf{x}_i; \hat\Theta_e) $.


\paragraph{Determine the number of latent factors.}
Before training AEALT, we must choose a critical hyperparameter: the number of neurons (or factors) in the bottleneck layer, which is the narrowest layer of the supervised autoencoder and defines the dimension of the latent representation $\mathbf{f}_i^* = \rho(\mathbf{x}_i; \Theta_e)$. In traditional linear factor models, the number of factors has a clear interpretation, and the methods for determining the number of factors in linear factor models have been extensively studied in the literature \citep{bai2002determining,hallin2007determining,onatski2009testing,kapetanios2010testing,onatski2010determining,lam2012factor,ahn2013eigenvalue,li2017determining,fan2022estimating}. However, in nonlinear factor models, this concept becomes ambiguous and is intertwined with the smoothness of the loading functions \citep{schmidt2021kolmogorov}. This challenge carries over to AEALT as well. In practice, we treat the number of factors as an architectural hyperparameter of the supervised autoencoder and select it through a model selection procedure.

\paragraph{Connections between AEALT and other post-processing approaches for text embeddings.}
Both PCA and the standard autoencoder emerge as special cases of the proposed AEALT framework. PCA can be viewed as a single-layer linear mapping that omits the supervised prediction component, while the standard autoencoder arises by removing the prediction network from our supervised framework.

\subsection{Downstream Learning based on Latent Factors}

By reducing the dimensionality of LLM embeddings, we obtain lower-dimensional representations (or factors) that reduce computational burden and boost statistical efficiency in downstream tasks. These low-dimensional factors $\hat\bff_i$ are then fed into generic neural network models or classical models $\mathcal{H}(\cdot;\Theta_t)$ for applications such as text classification and regression. In our experiments, we address both standard multi-class classification and anomaly detection under severe class imbalance. Unlike the universal losses used in the latent factor extraction process, the loss functions for fitting $\mathcal{H}(\cdot;\Theta_t)$ are problem- and algorithm-specific; for example, SVM uses hinge loss.

\section{Evaluation} \label{sec:evaluation}

To systematically study the effect of dimension reduction on LLM embeddings for downstream tasks as well as demonstrate the effectiveness of our proposed AEALT method, we conduct experiments on a variety of real-world text datasets spanning classification, anomaly detection, and prediction tasks.
For all learning tasks, we compare the proposed AEALT\footnote{We will make code publicly available upon acceptance.} to three baseline approaches: dimension reduction of embeddings using Principal Component Analysis (PCA) and a standard Autoencoder (AE), as well as a Vanilla approach that directly utilizes high-dimensional features with no dimension reduction. For downstream analysis, we employ various learning methods. For clarity, we adopt a naming convention: (DimensionReductionMethod)-(LearningMethod),  e.g., AEALT-Method, AE-Method, PCA-Method, and Vanilla-Method, where ``Method'' depends on the learning task. For all the datasets, we use 70\% for training and 30\% for testing. 

\subsection{Sentiment Analysis}

Sentiment analysis has become an increasingly valuable tool in financial applications, enabling the extraction of opinions, emotions, and attitudes from textual data such as news articles, analyst reports, and social media posts. By quantifying market sentiment, this technique helps investors, traders, and analysts assess public perception and forecast market movements more effectively, offering a data-driven complement to traditional financial indicators.
In this task, we assess the classification performance of various approaches in sentiment prediction. 

\bigskip 
\noindent\textbf{Datasets. } \quad 

We use five datasets for evaluation in this task: \texttt{FinEntity} \citep{tang-etal-2023-finentity} and four datasets from the Financial Phrase Bank \citep{Malo2014GoodDO}, including \texttt{Sentences\_50}, \texttt{Sentences\_66}, \texttt{Sentences\_75}, \texttt{Sentences\_100}. \\
FinEntity is an entity-level sentiment classification dataset built from financial news, containing labeled samples with positive, neutral, or negative sentiment.
Financial Phrase Bank is a widely used financial sentiment corpus, partitioned into four subsets based on the annotator agreement levels, with annotations provided by multiple human experts. 
Each subset corresponds to the minimum level of agreement among annotators. For example, \texttt{Sentences\_75} includes only sentences for which at least 75\% of the experts agreed on the sentiment label. Higher agreement levels indicate more distinct sentiment polarity. The five datasets range from 1,600 to 5,000 labeled samples.

\bigskip 
\noindent\textbf{Evaluation Protocol. } \quad 

We aim to assess the classification performance of different methods across various datasets. 
For the embedding model, we employ the FinBERT \citep{huang2022finbert}, which is 
specifically pre-trained on financial corpora, to obtain domain-specific sentence embeddings. 
For evaluation metrics, we consider Accuracy and F1 score. Each experiment is repeated 20 times. The evaluation spans both traditional classifiers (Logistic Regression~\citep{cox1958regression}, SVM~\citep{cortes1995support}, MLP~\citep{rosenblatt1958perceptron}) and advanced gradient-boosted tree methods such as LightGBM~\citep{ke2017lightgbm} and XGBoost~\citep{chen2016xgboost} as well as state-of-art algorithm such as Structured State Space (S4) \citep{gu2022efficiently} and RFE-GRU\citep{shams2025rfegru}. Results are summarized in Table~\ref{tab:classification-results}.

\bigskip
\noindent\textbf{Empirical Results. } \quad

As shown in Table~\ref{tab:classification-results}, the proposed AEALT consistently achieves both the best and the second-best performance across all datasets. Comparing dimension reduction methods to the Vanilla approach, PCA incurs a slight loss of accuracy but remains broadly competitive with Vanilla. This outcome aligns with findings from other PCA-based text learning studies in the literature \citep{zhang2024evaluating} and can be interpreted as a natural result of the trade-off between computational cost and algorithm performance. AE offers a limited gain over PCA at the expense of increased computational cost, highlighting the difficulty of extracting meaningful low-dimensional features from text embeddings. 
In contrast, AEALT consistently delivers substantial performance improvements of 5\%--15\% over Vanilla across all classification algorithms. These results demonstrate that AEALT, benefiting from its supervised design, effectively extracts task-relevant low-dimensional latent representations from text embeddings, thereby improving the overall predictive performance of downstream models.

We also note that in \texttt{Sentences\_75} and  \texttt{Sentences\_100} where sentiment tendencies are particularly strong and explicit, most classification models already achieve Accuracy and F1 scores above 0.9. Therefore, even a 5\% relative improvement by AEALT is significant and non-trivial.
In contrast, datasets \texttt{Sentences\_50} and \texttt{Sentences\_66} represent more subtle and nuanced sentiment expressions. On these more ambiguous datasets, AEALT yields around 10\% improvement in both Accuracy and F1 score. This demonstrates the strength of our supervised AEALT method in capturing fine-grained linguistic cues and subtle emotional undertones in the input text, enabling the construction of more effective low-dimensional representations that enhance model performance.

\begin{table*}[htbp]
\centering
\caption{Comparison of performance in sentiment analysis, evaluated using Accuracy and F1 score, across various methods and datasets. Results are averaged over 20 repetitions. \textbf{\textcolor{red}{Red}} highlights the best score for each metric in every column, while \textbf{\textcolor{blue}{blue}} marks the second-best.}
\label{tab:classification-results}
\setlength{\tabcolsep}{4pt}
\resizebox{\textwidth}{!}{%
\begin{tabular}{l|cc|cc|cc|cc|cc}
\toprule
& \multicolumn{2}{c|}{\textbf{FinEntity}} & \multicolumn{2}{c|}{\textbf{Sentences\_50}} & \multicolumn{2}{c|}{\textbf{Sentences\_66}} & \multicolumn{2}{c|}{\textbf{Sentences\_75}} & \multicolumn{2}{c}{\textbf{Sentences\_100}} \\
& Accuracy & F1 & Accuracy & F1 & Accuracy & F1 & Accuracy & F1 & Accuracy & F1 \\

\midrule
AEALT-Logistic & 0.874 & 0.858 & \textbf{\textcolor{blue}{0.936}} & \textbf{\textcolor{blue}{0.926}} & \textbf{\textcolor{blue}{0.948}} & \textbf{\textcolor{blue}{0.937}} & 0.963 & 0.955 & 0.977 & 0.965 \\
AE-Logistic & 0.799 & 0.780 & 0.830 & 0.800 & 0.871 & 0.844 & 0.916 & 0.890 & 0.954 & 0.936 \\
PCA-Logistic & 0.769 & 0.743 & 0.824 & 0.793 & 0.869 & 0.848 & 0.907 & 0.881 & 0.958 & 0.944 \\
Vanilla-Logistic & 0.765 & 0.743 & 0.796 & 0.773 & 0.853 & 0.831 & 0.913 & 0.887 & 0.966 & 0.957 \\
\midrule
AEALT-SVM & \textbf{\textcolor{red}{0.892}} & \textbf{\textcolor{red}{0.881}} & \textbf{\textcolor{red}{0.937}} & \textbf{\textcolor{red}{0.927}} & 0.943 & 0.931 & \textbf{\textcolor{red}{0.967}} & \textbf{\textcolor{red}{0.960}} & \textbf{\textcolor{red}{0.987}} & \textbf{\textcolor{red}{0.979}} \\
AE-SVM & 0.807 & 0.784 & 0.855 & 0.824 & 0.895 & 0.875 & 0.920 & 0.896 & 0.965 & 0.945 \\
PCA-SVM & 0.796 & 0.768 & 0.823 & 0.782 & 0.867 & 0.838 & 0.911 & 0.884 & 0.959 & 0.946 \\
Vanilla-SVM & 0.799 & 0.772 & 0.838 & 0.810 & 0.887 & 0.867 & 0.918 & 0.890 & 0.965 & 0.952 \\
\midrule
AEALT-MLP & \textbf{\textcolor{blue}{0.885}} & \textbf{\textcolor{blue}{0.871}} & 0.933 & 0.923 & \textbf{\textcolor{red}{0.949}} & \textbf{\textcolor{red}{0.940}} & 0.964 & 0.955 & \textbf{\textcolor{blue}{0.982}} & \textbf{\textcolor{blue}{0.973}} \\
AE-MLP & 0.797 & 0.775 & 0.846 & 0.819 & 0.887 & 0.867 & 0.916 & 0.892 & 0.963 & 0.945 \\
PCA-MLP & 0.769 & 0.743 & 0.826 & 0.793 & 0.869 & 0.846 & 0.907 & 0.880 & 0.958 & 0.945 \\
Vanilla-MLP & 0.779 & 0.758 & 0.810 & 0.785 & 0.867 & 0.840 & 0.914 & 0.889 & 0.959 & 0.944 \\
\midrule
AEALT-LightGBM & 0.878 & 0.863 & 0.933 & 0.923 & 0.940 & 0.929 & 0.963 & 0.954 & \textbf{\textcolor{blue}{0.982}} & 0.971 \\
AE-LightGBM & 0.799 & 0.775 & 0.851 & 0.824 & 0.888 & 0.867 & 0.914 & 0.889 & 0.963 & 0.945 \\
PCA-LightGBM & 0.761 & 0.731 & 0.815 & 0.779 & 0.865 & 0.840 & 0.906 & 0.879 & 0.957 & 0.943 \\
Vanilla-LightGBM & 0.786 & 0.755 & 0.837 & 0.809 & 0.881 & 0.857 & 0.918 & 0.892 & 0.959 & 0.947 \\
\midrule
AEALT-XGBoost & 0.872 & 0.856 & 0.924 & 0.913 & 0.945 & 0.935 & \textbf{\textcolor{blue}{0.966}} & \textbf{\textcolor{blue}{0.959}} & \textbf{\textcolor{blue}{0.982}} & \textbf{\textcolor{blue}{0.973}} \\
AE-XGBoost & 0.798 & 0.778 & 0.844 & 0.816 & 0.881 & 0.860 & 0.906 & 0.880 & 0.953 & 0.932 \\
PCA-XGBoost & 0.774 & 0.742 & 0.817 & 0.784 & 0.862 & 0.837 & 0.905 & 0.878 & 0.957 & 0.944 \\
Vanilla-XGBoost & 0.776 & 0.745 & 0.834 & 0.803 & 0.887 & 0.867 & 0.925 & 0.903 & 0.954 & 0.938 \\
\midrule
AEALT-RFE-GRU   & 0.872 & 0.857 & 0.918 & 0.904 & 0.934 & 0.922 & 0.954 & 0.943 & 0.969 & 0.953 \\
AE-RFE-GRU      & 0.735 & 0.706 & 0.795 & 0.756 & 0.834 & 0.797 & 0.863 & 0.828 & 0.912 & 0.884 \\
PCA-RFE-GRU     & 0.778 & 0.750 & 0.822 & 0.789 & 0.857 & 0.832 & 0.903 & 0.875 & 0.955 & 0.939 \\
Vanilla-RFE-GRU & 0.727 & 0.696 & 0.782 & 0.736 & 0.809 & 0.773 & 0.858 & 0.824 & 0.905 & 0.878 \\
\midrule
AEALT-S4   & 0.874 & 0.859 & 0.911 & 0.899 & 0.941 & 0.928 & 0.956 & 0.945 & 0.980 & 0.969 \\
AE-S4      & 0.786 & 0.764 & 0.827 & 0.803 & 0.873 & 0.848 & 0.910 & 0.884 & 0.961 & 0.941 \\
PCA-S4     & 0.754 & 0.728 & 0.814 & 0.782 & 0.855 & 0.827 & 0.892 & 0.860 & 0.946 & 0.928 \\
Vanilla-S4 & 0.775 & 0.752 & 0.820 & 0.790 & 0.876 & 0.857 & 0.916 & 0.891 & 0.964 & 0.953 \\
\bottomrule
\end{tabular}
}
\end{table*}


\subsection{Anomaly Detection} \label{subsec:anomaly-detection}

Anomaly detection in Natural Language Processing (NLP) involves identifying text data that deviates significantly from the norm, which may indicate errors, unusual patterns, or potentially malicious content. This can include detecting outliers such as fraudulent reviews, unexpected shifts in sentiment, rare linguistic structures, or abnormal communication patterns in chat logs or financial reports. By learning the typical distribution of language features, such as syntax, semantics, or topic distributions, NLP models can flag instances that diverge from expected behavior. Anomaly detection plays a critical role in applications like fraud detection, content moderation, cybersecurity, and monitoring financial disclosures for irregularities. In this task, we assess the imbalanced classification performance of various approaches in anomaly detection. 

\bigskip 
\noindent\textbf{Datasets.} \quad

We use four datasets from the \texttt{NLP\_by\_BERT} module of ADBench~\citep{han2022adbench}  for evaluation in this task: , namely: \texttt{20News}, \texttt{AGNews}, \texttt{IMDB}, and \texttt{YELP}. These datasets are collected from various well-known NLP benchmarks and curated for anomaly detection evaluation. Each dataset contains over 10,000 samples and maintains an approximate anomaly ratio of 5\%, reflecting realistic imbalanced detection settings.

\bigskip 
\noindent\textbf{Evaluation Protocol.} \quad

We aim to assess the performance of different methods across various datasets in this imbalanced classification task.  
For the embedding model, all text samples are pre-embedded using a pre-trained BERT \citep{devlin-etal-2019-bert} model provided by the benchmark suite, allowing direct application of downstream models. 
For evaluation metrics, following ADBench’s protocol, we use F1 score, AUROC (Area Under the Receiver Operating Characteristic Curve), and AUCPR (Area Under the Precision-Recall Curve) as the main evaluation metrics to account for class imbalance.
We also report the Precision, Recall, and Accuracy.
Due to the highly imbalanced nature of anomaly detection tasks, the classifier threshold is determined in a data-driven manner by optimizing the F1 score on the training set.
Each experiment is repeated 20 times.
The approaches include the unsupervised Isolation Forest (IForest)~\citep{liu2012isolation}, the semi-supervised DeepSAD~\citep{ruff2020deepsemisupervisedanomalydetection}, the supervised SVM~\citep{cortes1995support}, MLP~\citep{rosenblatt1958perceptron}, LightGBM~\citep{ke2017lightgbm},
XGBoost~\citep{chen2016xgboost}, and Structured State Space (S4) method~\citep{gu2022efficiently}. Results are summarized in Table~\ref{tab:anomaly-detection-results} and Table~\ref{tab:anomaly-rest}.

\bigskip 
\noindent\textbf{Empirical Results. } \quad

As shown in Table~\ref{tab:anomaly-detection-results}, AEALT achieves either the best or second-best F1 score and AUCPR on all four datasets, demonstrating its precise detection capability in the presence of rare anomalies. As for AUROC, AEALT also attains the best performance on \texttt{IMDB} and \texttt{YELP}, but underperforms the Vanilla baseline on \texttt{20NEWs} and \texttt{AGNEWs}, possibly due to the highly imbalanced nature of anomaly detection tasks.
In this task,
AE exhibits highly variable performance across datasets and algorithms. For instance, while AE-SVM achieves a sizeable F1 improvement on \texttt{IMDB}, AE-S4 suffers a 40\% drop. On average, AE leads to a 10\%--25\% performance degradation across all metrics. PCA performs even worse, with F1 scores and AUCPR typically reduced by 30\%--60\%, and AUROC dropping by 10\%--20\% across datasets. These results reveal the limitations of traditional unsupervised dimension reduction methods in capturing meaningful anomaly-relevant features, which constitute only a small fraction of the data.
AEALT shows consistent improvements over the Vanilla baseline in F1 score and AUCPR, with improvements ranging from approximately 20\% to as high as 50\% and MLP being the only exception. The improvement in AUROC is less substantial compared to F1, yet the supervised nature of AEALT still provides a noticeable benefit. For instance, when paired with the unsupervised IForest model, AEALT contributes to a relative AUROC gain of approximately 10\% - 20\%. 

In addition, we also report Recall, Precision, and Accuracy in Table \ref{tab:anomaly-rest}.
We acknowledge that, due to the highly imbalanced nature of the samples in the anomaly detection task, Accuracy offers limited interpretability and should not be overemphasized. 
Additionally, either Precision or Recall alone may fail to provide a reliable assessment of model performance in classification on highly imbalanced datasets.
We observe that many methods tend to achieve a high score in one metric while performing poorly in the other. For example, the PCA-SVM achieves perfect Recall (1.0) on the \texttt{IMDB} and \texttt{YELP} datasets, but this comes at the cost of extremely low Precision, reaching only 0.05. In contrast, the proposed AEALT framework appears to strike a good balance between Recall and Precision across most cases, demonstrating its effectiveness as a practical anomaly detection solution.


Our findings are globally consistent with those reported in ADBench~\citep{han2022adbench}: unsupervised and semi-supervised methods generally underperform compared to supervised approaches, often achieving only around half the performance in terms of F1 and AUCPR, or even less. We further extend this observation to the dimension reduction stage, as unsupervised methods like AE and PCA also exhibit poor performance in this context compared with Vanilla or supervised AEALT. Moreover, applying supervised dimension reduction significantly enhances the effectiveness of unsupervised anomaly detectors, further demonstrating the effectiveness of the proposed AEALT framework.

\begin{table*}[htbp]
\centering
\caption{Comparison of anomaly detection performance, evaluated using F1 score, AUROC and AUCPR, across a range of models and benchmark datasets. Results are averaged over 20 repetitions. \textbf{\textcolor{red}{Red}} highlights the best-performing model for each metric in a given column, while \textbf{\textcolor{blue}{blue}} indicates the second-best.
}
\label{tab:anomaly-detection-results}
\resizebox{\textwidth}{!}{%
\begin{tabular}{l|ccc|ccc|ccc|ccc}
\toprule
 & \multicolumn{3}{c|}{\textbf{20NEWS}} & \multicolumn{3}{c|}{\textbf{AGNEWS}} & \multicolumn{3}{c|}{\textbf{IMDB}} & \multicolumn{3}{c}{\textbf{YELP}} \\
 & F1 & AUROC & AUCPR & F1 & AUROC & AUCPR & F1 & AUROC & AUCPR & F1 & AUROC & AUCPR \\
\midrule
AEALT-IForest     & 0.288 & 0.753 & 0.211 & 0.305 & 0.774 & 0.205 & 0.148 & 0.625 & 0.076 & 0.213 & 0.704 & 0.136 \\
AE-IForest\        & 0.111 & 0.563 & 0.070 & 0.117 & 0.588 & 0.070 & 0.099 & 0.528 & 0.051 & 0.120 & 0.604 & 0.070 \\
PCA-IForest       & 0.089 & 0.554 & 0.063 & 0.108 & 0.544 & 0.066 & 0.095 & 0.508 & 0.048 & 0.101 & 0.557 & 0.061 \\
Vanilla-IForest  & 0.109 & 0.565 & 0.063 & 0.116 & 0.595 & 0.070 & 0.095 & 0.512 & 0.049 & 0.108 & 0.588 & 0.066 \\
\midrule
AEALT-DeepSAD     & 0.229 & 0.733 & 0.230 & 0.301 & 0.791 & 0.327 & 0.102 & 0.564 & 0.067 & 0.233 & 0.702 & 0.164 \\
AE-DeepSAD        & 0.120 & 0.587 & 0.080 & 0.159 & 0.650 & 0.097 & 0.097 & 0.532 & 0.052 & 0.132 & 0.628 & 0.081 \\
PCA-DeepSAD      & 0.102 & 0.583 & 0.072 & 0.117 & 0.576 & 0.070 & 0.096 & 0.503 & 0.049 & 0.113 & 0.582 & 0.073 \\
Vanilla-DeepSAD   & 0.137 & 0.669 & 0.114 & 0.228 & 0.744 & 0.174 & 0.092 & 0.517 & 0.051 & 0.151 & 0.653 & 0.091 \\
\midrule
AEALT-SVM         & 0.480 & 0.858 & \textbf{\textcolor{red}{0.477}} & \textbf{\textcolor{red}{0.673}} & 0.928 & \textbf{\textcolor{red}{0.698}} & \textbf{\textcolor{red}{0.483}} & 0.883 & \textbf{\textcolor{red}{0.471}} & \textbf{\textcolor{blue}{0.540}} & 0.905 & \textbf{\textcolor{red}{0.558}} \\
AE-SVM       & 0.185 & 0.675 & 0.165 & 0.481 & 0.876 & 0.484 & 0.176 & 0.719 & 0.146 & 0.314 & 0.820 & 0.252 \\
PCA-SVM          & 0.121 & 0.620 & 0.115 & 0.160 & 0.700 & 0.158 & 0.095 & 0.330 & 0.034 & 0.095 & 0.660 & 0.153 \\
Vanilla-SVM       & 0.241 & 0.831 & 0.249 & 0.515 & 0.922 & 0.547 & 0.048 & 0.833 & 0.201 & 0.396 & 0.876 & 0.367 \\
\midrule
AEALT-MLP        & 0.466 & 0.871 & 0.448 & 0.653 & 0.928 & 0.675 & 0.425 & 0.880 & 0.387 & 0.517 & 0.910 & 0.493 \\
AE-MLP            & 0.252 & 0.770 & 0.221 & 0.513 & 0.904 & 0.544 & 0.283 & 0.802 & 0.234 & 0.381 & 0.867 & 0.341 \\
PCA-MLP          & 0.157 & 0.702 & 0.134 & 0.263 & 0.771 & 0.220 & 0.123 & 0.692 & 0.095 & 0.154 & 0.691 & 0.110 \\
Vanilla-MLP      & 0.465 & \textbf{\textcolor{blue}{0.881}} & 0.408 & 0.654 & \textbf{\textcolor{red}{0.947}} & \textbf{\textcolor{red}{0.698}} & 0.442 & 0.876 & 0.419 & 0.534 & 0.912 & 0.541 \\
\midrule
AEALT-LightGBM   & 0.454 & 0.854 & 0.453 & 0.659 & 0.944 & \textbf{\textcolor{red}{0.698}} & 0.427 & \textbf{\textcolor{red}{0.894}} & \textbf{\textcolor{blue}{0.452}} & \textbf{\textcolor{red}{0.544}} & \textbf{\textcolor{red}{0.924}} & \textbf{\textcolor{blue}{0.545}} \\
AE-LightGBM      & 0.199 & 0.748 & 0.197 & 0.480 & 0.902 & 0.502 & 0.208 & 0.808 & 0.221 & 0.263 & 0.847 & 0.261 \\
PCA-LightGBM     & 0.160 & 0.702 & 0.147 & 0.170 & 0.788 & 0.206 & 0.103 & 0.731 & 0.115 & 0.110 & 0.746 & 0.136 \\
Vanilla-LightGBM & 0.217 & 0.844 & 0.303 & 0.571 & 0.932 & 0.620 & 0.400 & 0.869 & 0.390 & 0.450 & \textbf{\textcolor{blue}{0.916}} & 0.435 \\
\midrule
AEALT-XGBoost    & \textbf{\textcolor{blue}{0.480}} & 0.849 & \textbf{\textcolor{blue}{0.454}} & 0.658 & 0.938 & \textbf{\textcolor{blue}{0.683}} & \textbf{\textcolor{blue}{0.448}} & \textbf{\textcolor{blue}{0.887}} & 0.428 & 0.536 & 0.909 & 0.529 \\
AE-XGBoost       & 0.223 & 0.752 & 0.192 & 0.491 & 0.901 & 0.498 & 0.252 & 0.801 & 0.209 & 0.333 & 0.848 & 0.274 \\
PCA-XGBoost      & 0.146 & 0.706 & 0.148 & 0.220 & 0.778 & 0.196 & 0.111 & 0.713 & 0.106 & 0.147 & 0.729 & 0.124 \\
Vanilla-XGBoost  & 0.315 & 0.837 & 0.317 & 0.570 & 0.932 & 0.601 & 0.351 & 0.860 & 0.303 & 0.466 & 0.911 & 0.442 \\
\midrule
AEALT-S4         & \textbf{\textcolor{red}{0.481}} & 0.870 & 0.451 & \textbf{\textcolor{blue}{0.661}} & 0.930 & 0.672 & 0.433 & 0.879 & 0.393 & 0.539 & 0.909 & 0.521 \\
AE-S4            & 0.260 & 0.778 & 0.221 & 0.531 & 0.903 & 0.556 & 0.240 & 0.784 & 0.202 & 0.324 & 0.842 & 0.280 \\
PCA-S4           & 0.182 & 0.703 & 0.137 & 0.267 & 0.781 & 0.227 & 0.146 & 0.716 & 0.116 & 0.163 & 0.708 & 0.132 \\
Vanilla-S4       & 0.470 & \textbf{\textcolor{red}{0.891}} & 0.428 & 0.656 & \textbf{\textcolor{blue}{0.945}} & \textbf{\textcolor{blue}{0.683}} & 0.400 & 0.861 & 0.376 & 0.524 & 0.907 & 0.521 \\
\bottomrule
\end{tabular}
}

\end{table*}

\begin{table*}
\centering
\caption{Comparison of anomaly detection performance across various models and input representations, evaluated using Accuracy, Recall, and Precision. Results are averaged over 20 repetitions. \textbf{\textcolor{red}{Red}} denotes the best-performing model for each metric, while \textbf{\textcolor{blue}{blue}} indicates the second-best.
}
\label{tab:anomaly-rest}
\setlength{\tabcolsep}{3pt}
\resizebox{\textwidth}{!}{%
\begin{tabular}{l|ccc|ccc|ccc|ccc}
\toprule
& \multicolumn{3}{c}{\textbf{20NEWS}} & \multicolumn{3}{c}{\textbf{AGNEWS}} & \multicolumn{3}{c}{\textbf{IMDB}} & \multicolumn{3}{c}{\textbf{YELP}} \\
\cmidrule(lr){2-4} \cmidrule(lr){5-7} \cmidrule(lr){8-10} \cmidrule(lr){11-13}
& Accuracy & Recall & Precision & Accuracy & Recall & Precision & Accuracy & Recall & Precision & Accuracy & Recall & Precision \\
\midrule
AEALT-IForest      & 0.803 & 0.485 & 0.247 & 0.785 & 0.597 & 0.235 & 0.568 & 0.591 & 0.093 & 0.751 & 0.500 & 0.152 \\
AE-IForest         & 0.538 & 0.521 & 0.075 & 0.611 & 0.480 & 0.072 & 0.159 & 0.922 & 0.052 & 0.693 & 0.420 & 0.072 \\
PCA-IForest        & 0.628 & 0.409 & 0.054 & 0.574 & 0.491 & 0.067 & 0.117 & \textbf{\textcolor{blue}{0.927}} & 0.050 & 0.526 & 0.530 & 0.056 \\
Vanilla-IForest\    & 0.530 & \textbf{\textcolor{blue}{0.581}} & 0.064 & 0.587 & 0.522 & 0.066 & 0.050 & \textbf{\textcolor{red}{1.000}} & 0.050 & 0.620 & 0.460 & 0.061 \\
\midrule
AEALT-DeepSAD      & 0.800 & 0.364 & 0.308 & 0.759 & 0.496 & 0.356 & 0.384 & 0.659 & 0.068 & 0.878 & 0.322 & 0.225 \\
AE-DeepSAD         & 0.647 & 0.420 & 0.080 & 0.806 & 0.329 & 0.113 & 0.152 & 0.904 & 0.051 & 0.716 & 0.428 & 0.079 \\
PCA-DeepSAD        & 0.716 & 0.326 & 0.067 & 0.688 & 0.392 & 0.074 & 0.137 & 0.913 & 0.051 & 0.734 & 0.339 & 0.069 \\
Vanilla-DeepSAD    & 0.514 & \textbf{\textcolor{red}{0.612}} & 0.088 & 0.889 & 0.322 & 0.183 & 0.595 & 0.411 & 0.053 & 0.816 & 0.327 & 0.099 \\
\midrule
AEALT-MLP          & 0.945 & 0.470 & 0.486 & 0.965 & 0.655 & 0.654 & 0.940 & 0.442 & 0.415 & 0.952 & 0.508 & 0.526 \\
AE-MLP             & 0.919 & 0.271 & 0.244 & 0.945 & 0.569 & 0.469 & 0.916 & 0.333 & 0.246 & 0.929 & 0.442 & 0.337 \\
PCA-MLP            & 0.903 & 0.187 & 0.143 & 0.926 & 0.268 & 0.262 & 0.906 & 0.132 & 0.116 & 0.918 & 0.150 & 0.159 \\
Vanilla-MLP        & 0.938 & 0.528 & 0.466 & 0.965 & 0.677 & 0.638 & 0.936 & 0.507 & 0.392 & 0.950 & \textbf{\textcolor{blue}{0.573}} & 0.500 \\
\midrule
AEALT-SVM     & 0.944 & 0.460 & \textbf{\textcolor{red}{0.533}} & \textbf{\textcolor{red}{0.967}} & 0.671 & \textbf{\textcolor{blue}{0.678}} & \textbf{\textcolor{blue}{0.948}} & 0.489 & \textbf{\textcolor{blue}{0.484}} & \textbf{\textcolor{red}{0.956}} & 0.522 & 0.560 \\
AE-SVM        & 0.721 & 0.423 & 0.181 & 0.933 & 0.600 & 0.410 & 0.785 & 0.353 & 0.188 & 0.906 & 0.441 & 0.253 \\
PCA-SVM       & 0.778 & 0.280 & 0.139 & 0.287 & \textbf{\textcolor{red}{0.818}} & 0.137 & 0.050 & \textbf{\textcolor{red}{1.000}} & 0.050 & 0.050 & \textbf{\textcolor{red}{1.000}} & 0.050 \\
Vanilla-SVM   & 0.889 & 0.415 & 0.198 & 0.928 & \textbf{\textcolor{blue}{0.738}} & 0.402 & 0.947 & 0.027 & 0.250 & 0.915 & 0.560 & 0.307 \\
\midrule
AEALT-LightGBM    & 0.943 & 0.482 & 0.443 & 0.964 & 0.695 & 0.629 & \textbf{\textcolor{red}{0.949}} & 0.390 & \textbf{\textcolor{red}{0.502}} & 0.954 & 0.551 & 0.539 \\
AE-LightGBM       & 0.929 & 0.181 & 0.283 & 0.952 & 0.463 & 0.510 & 0.944 & 0.150 & 0.359 & 0.939 & 0.226 & 0.329 \\
PCA-LightGBM      & 0.909 & 0.168 & 0.167 & 0.944 & 0.119 & 0.316 & 0.938 & 0.071 & 0.189 & 0.937 & 0.078 & 0.190 \\
LightGBM  & 0.947 & 0.170 & 0.483 & 0.949 & 0.673 & 0.497 & 0.938 & 0.413 & 0.388 & 0.936 & 0.527 & 0.393 \\
\midrule
AEALT-XGBoost     & \textbf{\textcolor{blue}{0.948}} & 0.479 & 0.493 & 0.964 & 0.678 & 0.641 & 0.944 & 0.460 & 0.444 & \textbf{\textcolor{blue}{0.953}} & 0.546 & 0.528 \\
AE-XGBoost       & 0.921 & 0.227 & 0.234 & 0.945 & 0.528 & 0.460 & 0.929 & 0.244 & 0.268 & 0.930 & 0.359 & 0.314 \\
PCA-XGBoost       & 0.903 & 0.166 & 0.134 & 0.935 & 0.188 & 0.275 & 0.928 & 0.091 & 0.146 & 0.930 & 0.122 & 0.189 \\
Vanilla-XGBoost   & 0.942 & 0.275 & 0.432 & 0.950 & 0.645 & 0.515 & 0.917 & 0.447 & 0.289 & 0.938 & 0.540 & 0.409 \\
\midrule
AEALT-S4          & \textbf{\textcolor{red}{0.950}} & 0.470 & \textbf{\textcolor{blue}{0.510}} & 0.967 & 0.654 & 0.673 & 0.945 & 0.424 & 0.444 & \textbf{\textcolor{red}{0.956}} & 0.520 & \textbf{\textcolor{blue}{0.562}} \\
AE-S4            & 0.931 & 0.244 & 0.294 & 0.954 & 0.531 & 0.541 & 0.929 & 0.226 & 0.264 & 0.932 & 0.332 & 0.321 \\
PCA-S4            & 0.919 & 0.186 & 0.198 & 0.933 & 0.253 & 0.289 & 0.913 & 0.149 & 0.145 & 0.923 & 0.151 & 0.179 \\
Vanilla-S4        & 0.946 & 0.496 & 0.480 & \textbf{\textcolor{blue}{0.967}} & 0.631 & \textbf{\textcolor{red}{0.687}} & 0.945 & 0.367 & 0.445 & \textbf{\textcolor{red}{0.956}} & 0.490 & \textbf{\textcolor{red}{0.567}} \\
\bottomrule
\end{tabular}
}
\end{table*}

\subsection{Price Prediction}

Price prediction using textual data involves analyzing unstructured information, such as financial news, social media posts, earnings calls, and analyst reports, to forecast future asset prices. The underlying idea is that market-relevant text often contains early signals of sentiment, events, or trends that can influence investor decisions and thus impact prices. In this task, we assess the predictive performances of various methods using product descriptions. 

\bigskip 
\noindent\textbf{Datasets.} \quad

We use the Amazon Product Sales Dataset 2023~\citep{rao2023amazon}, which spans more than one hundred product categories. To ensure adequate statistical power, we focus on five categories with the largest sample sizes: \texttt{Appliances}, \texttt{Entertainment}, \texttt{Heating},  \texttt{Kitchen}, and  \texttt{Sportswear},each containing over 7,500 samples. The target variable \(y\) (price) is standardized prior to model training.

\bigskip 
\noindent\textbf{Evaluation Protocol.} \quad

We aim to assess the predictive performance of different methods across various datasets. For the embedding model, we use a pre-trained BERT \citep{devlin-etal-2019-bert} to obtain embeddings for each product description.
For evaluation metrics, performance is assessed using Mean Absolute Error (MAE), Root Mean Squared Error (RMSE), and Out-of-Sample \(R^2\) (\(R^2_{\text{oos}}\)).  Each experiment is repeated 20 times.  We conduct a comprehensive evaluation of several models:    LASSO~\citep{tibshirani2011retrospective}, MLP~\citep{rosenblatt1958perceptron}, XGBoost~\citep{chen2016xgboost},  LightGBM~\citep{ke2017lightgbm} and state-of-art algorithm such as RFE-GRU \citep{shams2025rfegru}, MLP-PLR \citep{NEURIPS2022_9e9f0ffc}, REAL-MLP \citep{NEURIPS2024_2ee1c872}.
Results are summarized in Table~\ref{tab:prediction-results}.

\bigskip
\noindent\textbf{Empirical Results. } \quad

As shown in Table~\ref{tab:prediction-results}, algorithms equipped with AEALT consistently achieve the best performance in MAE, RMSE and \(R_{\rm oos}^2\) across nearly all datasets, with the sole exception being the RMSE on the \texttt{Sportswear} dataset, showcasing its strength in extracting price-relevant signals from text embeddings. PCA underperforms the Vanilla baseline by a large margin in this talk, while AE shows improvement over PCA but still falls short of matching the Vanilla performance. Meanwhile,  AEALT yields an average performance gain of approximately 10\% over Vanilla across most settings.

\begin{table*}[htbp]
\centering
\caption{Comparison of prediction performance with textual data embeddings, evaluated using MAE, RMSE and \(R^2_{\text{oos}}\) across various models and benchmark datasets. Results are averaged over 20 repetitions. \textbf{\textcolor{red}{Red}} denotes the best-performing model for each metric, while \textbf{\textcolor{blue}{blue}} indicates the second-best.}
\label{tab:prediction-results}
\vspace{-2ex}
\setlength{\tabcolsep}{3pt}
\resizebox{\textwidth}{!}{%
\renewcommand{\arraystretch}{1.1}
\begin{tabular}{l|ccc|ccc|ccc|ccc|ccc}
\toprule
& \multicolumn{3}{c|}{\textbf{APPLIANCES}} & \multicolumn{3}{c|}{\textbf{ENTERTAINMENT}} & \multicolumn{3}{c|}{\textbf{HEATING}} & \multicolumn{3}{c|}{\textbf{KITCHEN}} & \multicolumn{3}{c}{\textbf{SPORTSWEAR}} \\
\cmidrule(lr){2-4} \cmidrule(lr){5-7} \cmidrule(lr){8-10} \cmidrule(lr){11-13} \cmidrule(lr){14-16}
& MAE & RMSE & \(R^2_{\text{oos}}\) & MAE & RMSE & \(R^2_{\text{oos}}\) & MAE & RMSE & \(R^2_{\text{oos}}\) & MAE & RMSE & \(R^2_{\text{oos}}\) & MAE & RMSE & \(R^2_{\text{oos}}\) \\
\midrule
AEALT-LASSO        & 0.374 & 0.656 & 0.614 & 0.271 & 0.611 & 0.625 & 0.348 & 0.527 & 0.671 & 0.381 & 0.627 & 0.608 & 0.521 & 0.748 & 0.485 \\
AE-LASSO           & 0.496 & 0.905 & 0.283 & 0.360 & 0.883 & 0.235 & 0.479 & 0.729 & 0.383 & 0.482 & 0.902 & 0.201 & 0.590 & 0.811 & 0.396 \\
PCA-LASSO          & 0.514 & 0.960 & 0.193 & 0.360 & 0.937 & 0.139 & 0.500 & 0.823 & 0.214 & 0.489 & 0.932 & 0.148 & 0.635 & 0.847 & 0.340 \\
Vanilla-LASSO      & 0.441 & 0.780 & 0.467 & 0.330 & 0.748 & 0.451 & 0.393 & 0.583 & 0.605 & 0.453 & 0.756 & 0.440 & 0.521 & 0.747 & 0.487 \\
\midrule
AEALT-MLP          & 0.261 & 0.548 & 0.736 & 0.159 & 0.486 & 0.767 & 0.233 & 0.434 & 0.781 & 0.314 & 0.587 & 0.662 & 0.481 & 0.767 & 0.459 \\
AE-MLP             & 0.351 & 0.687 & 0.586 & 0.197 & 0.571 & 0.680 & 0.334 & 0.543 & 0.657 & 0.461 & 0.759 & 0.434 & 0.601 & 0.892 & 0.268 \\
PCA-MLP            & 0.482 & 0.914 & 0.268 & 0.286 & 0.736 & 0.469 & 0.399 & 0.674 & 0.472 & 0.546 & 0.919 & 0.171 & 0.638 & 0.916 & 0.229 \\
Vanilla-MLP        & 0.311 & 0.561 & 0.724 & 0.232 & 0.527 & 0.728 & 0.294 & 0.466 & 0.748 & 0.386 & 0.629 & 0.612 & 0.493 & 0.761 & 0.467 \\
\midrule
AEALT-LightGBM          & \textcolor{red}{\textbf{0.232}} & \textcolor{blue}{\textbf{0.544}} & \textcolor{blue}{\textbf{0.741}} & \textcolor{blue}{\textbf{0.133}} & 0.500 & 0.755 & \textcolor{red}{\textbf{0.208}} & \textcolor{red}{\textbf{0.423}} & \textcolor{red}{\textbf{0.792}} & 0.272 & 0.561 & 0.691 & \textcolor{red}{\textbf{0.442}} & 0.722 & 0.521 \\
AE-LightGBM             & 0.334 & 0.699 & 0.572 & 0.213 & 0.617 & 0.627 & 0.287 & 0.528 & 0.676 & 0.370 & 0.704 & 0.514 & 0.494 & 0.740 & 0.497 \\
PCA-LightGBM            & 0.389 & 0.801 & 0.438 & 0.230 & 0.676 & 0.551 & 0.335 & 0.646 & 0.516 & 0.416 & 0.811 & 0.354 & 0.545 & 0.788 & 0.428 \\
Vanilla-LightGBM        & 0.306 & 0.658 & 0.622 & 0.206 & 0.595 & 0.653 & 0.269 & 0.513 & 0.695 & 0.334 & 0.624 & 0.618 & 0.461 & \textcolor{red}{\textbf{0.711}} & \textcolor{red}{\textbf{0.535}} \\
\midrule
AEALT-XGBoost     & \textcolor{blue}{\textbf{0.234}} & 0.549 & 0.735 & \textcolor{red}{\textbf{0.130}} & 0.512 & 0.742 & \textcolor{blue}{\textbf{0.213}} & 0.435 & 0.780 & 0.286 & 0.577 & 0.672 & 0.451 & 0.735 & 0.504 \\
AE-XGBoost        & 0.349 & 0.718 & 0.548 & 0.207 & 0.655 & 0.578 & 0.306 & 0.557 & 0.639 & 0.406 & 0.747 & 0.452 & 0.510 & 0.768 & 0.457 \\
PCA-XGBoost       & 0.407 & 0.820 & 0.411 & 0.230 & 0.722 & 0.488 & 0.360 & 0.678 & 0.466 & 0.448 & 0.852 & 0.288 & 0.561 & 0.819 & 0.383 \\
Vanilla-XGBoost   & 0.318 & 0.690 & 0.583 & 0.183 & 0.584 & 0.666 & 0.283 & 0.555 & 0.642 & 0.359 & 0.686 & 0.538 & 0.490 & 0.755 & 0.476 \\
\midrule
AEALT-RFE-GRU     & 0.286 & 0.577 & 0.708 & 0.197 & 0.556 & 0.697 & 0.246 & 0.437 & 0.778 & 0.324 & 0.589 & 0.659 & 0.464 & 0.735 & 0.503 \\
AE-RFE-GRU        & 0.401 & 0.773 & 0.476 & 0.274 & 0.696 & 0.524 & 0.343 & 0.565 & 0.629 & 0.422 & 0.771 & 0.416 & 0.519 & 0.752 & 0.481 \\
PCA-RFE-GRU       & 0.467 & 0.908 & 0.279 & 0.311 & 0.837 & 0.314 & 0.437 & 0.754 & 0.340 & 0.460 & 0.889 & 0.225 & 0.605 & 0.821 & 0.381 \\
Vanilla-RFE-GRU   & 0.350 & 0.613 & 0.671 & 0.281 & 0.579 & 0.671 & 0.293 & 0.468 & 0.746 & 0.397 & 0.627 & 0.615 & 0.481 & 0.721 & 0.522 \\
\midrule
AEALT-MLP-PLR     & 0.244 & 0.546 & 0.739 & 0.136 & 0.508 & 0.747 & 0.255 & 0.466 & 0.748 & \textcolor{blue}{\textbf{0.267}} & \textcolor{blue}{\textbf{0.551}} & \textcolor{blue}{\textbf{0.702}} & 0.468 & 0.740 & 0.496 \\
AE-MLP-PLR        & 0.383 & 0.726 & 0.539 & 0.197 & 0.732 & 0.475 & 0.508 & 0.702 & 0.428 & 0.331 & 0.641 & 0.597 & 0.513 & 0.746 & 0.489 \\
PCA-MLP-PLR       & 0.435 & 0.847 & 0.371 & 0.232 & 0.727 & 0.481 & 0.341 & 0.644 & 0.519 & 0.409 & 0.795 & 0.380 & 0.580 & 0.798 & 0.414 \\
Vanilla-MLP-PLR   & 0.294 & 0.610 & 0.675 & 0.147 & 0.500 & 0.755 & 0.235 & \textcolor{blue}{\textbf{0.427}} & \textcolor{blue}{\textbf{0.788}} & 0.321 & 0.601 & 0.646 & 0.504 & 0.726 & 0.515 \\
\midrule
AEALT-REAL-MLP    & 0.240 & \textcolor{red}{\textbf{0.538}} & \textcolor{red}{\textbf{0.747}} & 0.141 & \textcolor{red}{\textbf{0.450}} & \textcolor{red}{\textbf{0.802}} & 0.214 & 0.440 & 0.776 & \textcolor{red}{\textbf{0.267}} & \textcolor{red}{\textbf{0.539}} & \textcolor{red}{\textbf{0.715}} & \textcolor{blue}{\textbf{0.444}} & \textcolor{blue}{\textbf{0.719}} & \textcolor{blue}{\textbf{0.525}} \\
AE-REAL-MLP       & 0.336 & 0.702 & 0.569 & 0.175 & 0.592 & 0.656 & 0.265 & 0.481 & 0.732 & 0.350 & 0.665 & 0.566 & 0.508 & 0.746 & 0.489 \\
PCA-REAL-MLP      & 0.407 & 0.832 & 0.393 & 0.240 & 0.709 & 0.507 & 0.358 & 0.661 & 0.493 & 0.415 & 0.809 & 0.358 & 0.573 & 0.798 & 0.414 \\
Vanilla-REAL-MLP  & 0.271 & 0.581 & 0.705 & 0.152 & \textcolor{blue}{\textbf{0.475}} & \textcolor{blue}{\textbf{0.779}} & 0.237 & 0.433 & 0.783 & 0.297 & 0.574 & 0.676 & 0.487 & 0.739 & 0.498 \\
\bottomrule
\end{tabular}
}

\end{table*}

\section{Conclusion and Discussion} \label{sec:conclusion}
In this article, we propose AEALT, a unifying framework for supervised learning tasks with textual input. AEALT bridges unstructured text data and downstream tasks by extracting low-dimensional, task-relevant representations through a supervised dimension reduction process, implemented via a supervised autoencoder. 
We carry out extensive empirical studies using real-world datasets to systematically study how dimension reduction on LLM embeddings impacts downstream tasks. Using the Vanilla approach (where high-dimensional embeddings are directly fed into downstream tasks without any dimension reduction) as a baseline, empirical results in this study reveal the following: 
(i) PCA, an unsupervised method, often leads to performance degradation due to information loss during the dimension reduction process. While in some cases the performance drop is minor and PCA remains competitive with Vanilla, in others, its performance is significantly worse.
This aligns with prior findings in the literature \citep{zhang2024evaluating}, and can be interpreted as a trade-off between computational cost and algorithmic accuracy. 
(ii) AE, a DNN-based unsupervised dimension reduction method, performs better than PCA. However, its performance still often falls behind or is on par with Vanilla, suggesting that the lack of supervision limits its effectiveness in retaining task-specific information.
(iii) AEALT, the proposed DNN-based supervised approach, yields compelling advantages over the unsupervised PCA and AE, and frequently outperforms the Vanilla baseline. This demonstrates the effectiveness of AEALT in extracting task-relevant information from high-dimensional text embeddings, thereby enhancing the performance in downstream tasks.

{
\bibliographystyle{plainnat}
\bibliography{refs_AEALT}
 }

\end{document}